\documentclass[lettersize,journal]{IEEEtran}
\usepackage{amsmath,amsfonts}
\usepackage{algorithmic}
\usepackage{algorithm}
\usepackage{array}
\usepackage{graphicx}
\usepackage{subfigure} 
\usepackage{textcomp}
\usepackage{hyperref}
\usepackage{stfloats}
\usepackage{booktabs}
\usepackage{url}
\usepackage{verbatim}
\usepackage{multirow}
\usepackage{graphicx}
\usepackage{cite}
\usepackage{arydshln}
\usepackage{overpic}
\usepackage{color}
\usepackage{amssymb}
\usepackage{hyperref}

\hypersetup{
    colorlinks=true,
    linkcolor=blue,
    filecolor=blue,      
    urlcolor=blue,
    citecolor=blue,
}
\begin{document}

\title{Iterative Optimal Attention and Local Model for Single Image Rain Streak Removal}

\author{
    Xiangyu Li, Wanshu Fan, Yue Shen, Cong Wang, Wei Wang, Xin Yang, Qiang Zhang and Dongsheng Zhou

\thanks{This work was supported in part by the National Key Research and Development Program of China (Grant No. 2021ZD0112400), National Natural Science Foundation of China (Grant No. U1908214), the Program for Innovative Research Team in University of Liaoning Province (Grant No. LT2020015), the Support Plan for Key Field Innovation Team of Dalian (2021RT06), the Support Plan for Leading Innovation Team of Dalian University (XLJ202010), 111 Project (No. D23006), Interdisciplinary project of Dalian University (Grant No. DLUXK-2023-QN-015). (Corresponding author: Wanshu Fan, Dongsheng Zhou)}
\thanks{Xiangyu Li, Wanshu Fan and Yue Shen are with the National and Local Joint Engineering Laboratory of Computer Aided Design, School of Software Engineering, Dalian University, Dalian, China. (E-mail: lixiangyu@s.dlu.edu.cn; fanwanshu@dlu.edu.cn; chenyue@s.dlu.edu.cn).}
\thanks{Cong Wang is with the Department of Computing, The Hong Kong Polytechnic University, Hong Kong, China. (E-mail: supercong94@gmail.com).}
\thanks{
Wei Wang is with the School of Cyber Science and Technology, Sun Yat-Sen University, Shenzhen, China (E-mail: wangwei2@mail.sysu.edu.cn).}
\thanks{Xin Yang is with the School of Computer Science and Technology, Dalian University of Technology, Dalian, China. (E-mail: xinyang@dlut.edu.cn).}
\thanks{Qiang Zhang and Dongsheng Zhou are with the National and Local Joint Engineering Laboratory of Computer Aided Design, School of Software Engineering, Dalian University, School of Computer Science and Technology, Dalian University of Technology, Dalian, China. (E-mail: zhangq@dlu.edu.cn; zhouds@dlu.edu.cn).}
}
\date{\footnotesize\textsuperscript{\textbf{1}}National and Local Joint Engineering Laboratory of Computer Aided Design,\\
School of Software Engineering, Dalian University, Dalian 116622, LiaoNing, China\\ }
\markboth{Journal of \LaTeX\ Class Files,~Vol.~14, No.~8, August~2021}%
{Shell \MakeLowercase{\textit{et al.}}: A Sample Article Using IEEEtran.cls for IEEE Journals}


\maketitle

\begin{abstract}
High-fidelity imaging is crucial for the successful safety supervision and intelligent deployment of vision-based measurement systems (VBMS). 
It ensures high-quality imaging in VBMS, which is fundamental for reliable visual measurement and analysis. 
However, imaging quality can be significantly impaired by adverse weather conditions, particularly rain, leading to blurred images and reduced contrast. 
Such impairments increase the risk of inaccurate evaluations and misinterpretations in VBMS.
To address these limitations, we propose an Expectation Maximization Reconstruction Transformer (EMResformer) for single image rain streak removal. 
The EMResformer retains the key self-attention values for feature aggregation, enhancing local features to produce superior image reconstruction.
Specifically, we propose an Expectation Maximization Block seamlessly integrated into the single image rain streak removal network, enhancing its ability to eliminate superfluous information and restore a cleaner background image.
Additionally, to further enhance local information for improved detail rendition,
we introduce a Local Model Residual Block, which integrates two local model blocks along with a sequence of convolutions and activation functions. 
This integration synergistically facilitates the extraction of more pertinent features for enhanced single image rain streak removal. 
Extensive experiments validate that our proposed EMResformer surpasses current state-of-the-art single image rain streak removal methods on both synthetic and real-world datasets, achieving an improved balance between model complexity and single image deraining performance.
Furthermore, we evaluate the effectiveness of our method in VBMS scenarios, demonstrating that high-quality imaging significantly improves the accuracy and reliability of VBMS tasks.
Codes are at \href{https://github.com/ghfkahfk/EMResformer}{https://github.com/ghfkahfk/EMResformer}.
\end{abstract}

 
\section{Introduction}


\begin{figure}[!t] 
\centering
\includegraphics[width=\linewidth]{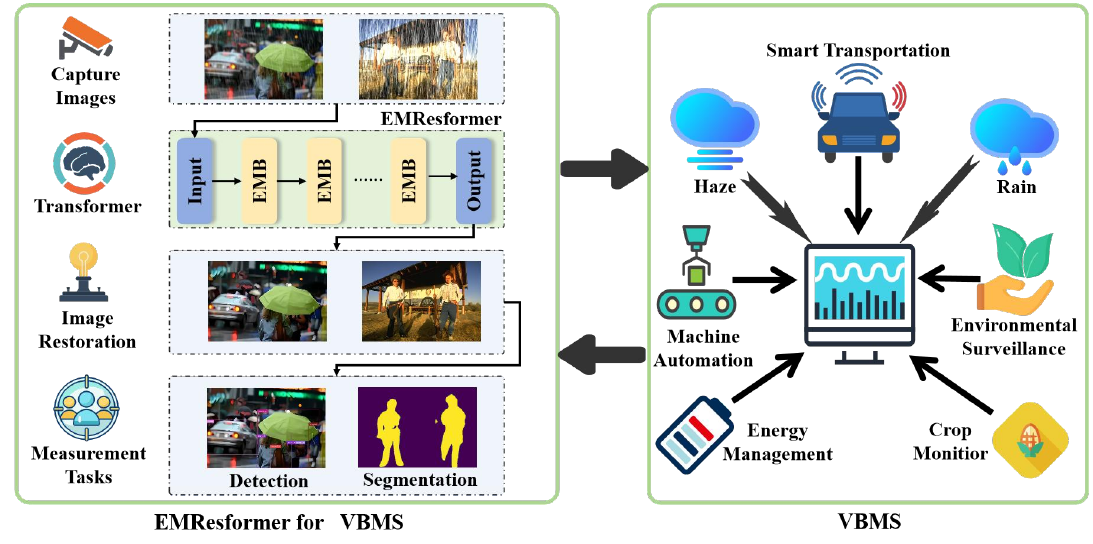} 
\vspace{-7mm}
\caption{Overview of our EMResformer for degraded scene restoration in vision-based measurement systems (VBMS).}
\label{fig:p7}
\vspace{-2mm}
\end{figure}

\IEEEPARstart{M}{achine}-vision and its related algorithms play a crucial role in instrumentation and measurement applications, forming key components of vision-based measurement systems (VBMS)~\cite{vbms1,celiang}. 
VBMS offer high accuracy and precision, making them widely utilized in areas such as smart transportation~\cite{transportation}, machine automation~\cite{automation}, environmental surveillance~\cite{environment}, energy management~\cite{batter}, crop monitior~\cite{food}, and other fields requiring precise measurements.
However, image quality degradation caused by adverse weather conditions, particularly rain, 
introducing uncertainties that impact VBMS repeatability.

As shown in Fig.\ref{fig:p7}, VBMS captures images, extracts features, and performs measurement tasks such as object detection and segmentation.
However, rainy conditions lead to image degradation, reducing feature extraction efficiency and measurement accuracy,
affecting the reliability of object detection and segmentation in VBMS.
This degradation leads to measurement inconsistencies, reduced precision, and increased uncertainty, which are critical concerns in instrumentation and measurement applications~\cite{0.4,0.2}. 
Studies in the instrumentation and measurement domain have shown that adverse weather conditions can introduce systematic and random errors in measurement systems, highlighting the importance of robust image restoration methods for ensuring stable and reliable VBMS performance.

The removal of rain streak plays a vital role in enhancing the performance of VBMS, including object detection~\cite{mubiaojiance1}, tracking~\cite{DBLP:conf/iclr/DosovitskiyB0WZ21}, and segmentation~\cite{fenge}. 
However, single image rain streak removal is inherently an ill-posed problem, as multiple plausible clean background images may correspond to the same rain-affected input image. This challenge underscores the necessity of effective rain removal techniques that not only restore visual clarity but also preserve measurement-critical features in VBMS.
Traditional approach~\cite{3} has attempted to address this issue by designing priors based on statistical characteristics of rain streaks and clear images.
However, these approaches struggle in complex and variable rainfall scenarios, limiting their effectiveness in precipitation attenuation.

In the context of VBMS, measurement accuracy and repeatability are key performance indicators, and they are often evaluated using well-established methodologies in the instrumentation and measurement field~\cite{vbms1,celiang}.  
Many existing computer vision-based rain removal methods focus solely on perceptual quality improvements rather than on their impact on measurement reliability in practical applications. 
Thus, there is a strong need for image restoration approaches specifically designed for VBMS that consider both visual enhancement and measurement stability under adverse conditions.

Recent years have witnessed significant advancements in deep learning, particularly in convolutional neural networks (CNNs)~\cite{8,9}. 
Leveraging extensive training datasets and sophisticated architectures, CNN-based methods excel in learning the mapping between rain-affected and clear images, surpassing traditional methods \cite{8,9}. 
However, CNNs~\cite{argan,8,9} have achieved significant success in inherent limitations, including static weights and limited receptive fields, which restrict their ability to model long-range dependencies critical for handling complex rain patterns.

In contrast, the attention mechanism of Transformers offers a global receptive field, enabling more comprehensive feature aggregation.
However, traditional attention mechanisms \cite{21,8} are computationally expensive, often generating large attention maps that may include redundant information, thereby reducing efficiency and effectiveness.
Furthermore, existing Transformer-based approaches often overlook local feature modeling, which is essential for preserving fine-grained measurement details in VBMS applications~\cite{0.4,0.2}.
Therefore, developing a Transformer-based framework that balances global feature aggregation with local detail preservation is crucial for achieving robust and accurate VBMS performance in adverse weather conditions.

To address these challenges, this paper proposes a novel framework that enhances image restoration for VBMS, ensuring improved feature extraction and measurement accuracy under complex conditions.
We  introduce EMResformer, a novel approach for single image rain streak removal that integrates the Expectation Maximization (EM) algorithm to iteratively optimize attention weights. This method reduces redundancy in feature maps and enhances the model's ability to recover fine image details, making it particularly suitable for vision-based measurement applications.
The Expectation Maximization Block (EMB) enhances feature attention by eliminating redundancy, thereby facilitating the restoration of clean images. 
Expectation Maximization Block (EMB) contain attention mechanisms that capture long-distance dependencies and lack the ability to model local information. 
So we propose a Local Model Residual Block (LMRB), which breaks down features into vertical and horizontal directions and aggregates them in different directions to get a more accurate representation.
These innovations result in significantly improved deraining performance, as demonstrated on synthetic datasets. 

The main contributions of this paper are summarised as follows:
\begin{itemize}
\item
We propose the Expectation-Maximization Reconstruction Transformer (EMResformer), which generates clearer feature maps, reduces redundancy, and improves the network's ability to accurately model rain streaks. By effectively removing rain streaks, EMResformer enhances the accuracy of vision-based tasks such as image segmentation and object detection. 
This is the first application of the expectation maximization attention mechanism for image rain streak removal.
%
\item 
We introduce the Expectation Maximization Block (EMB) and Local Model Residual Block (LMRB). 
The EMB iteratively optimizes attention weights to eliminate redundant information, improving the overall effectiveness of rain streak removal. 
The LMRB integrates local features and strengthens local modeling capabilities, improving image detail and texture preservation. Together, these components significantly enhance single image rain streak removal performance.
\item 
%
Extensive qualitative and quantitative experiments on widely-used benchmarks demonstrate that our EMResformer outperforms existing state-of-the-art rain streak removal methods. 
It also achieves superior segmentation performance and object detection accuracy in VBMS applications, demonstrating the effectiveness of the proposed rain streak removal algorithm and its potential to significantly enhance VBMS performance.

\end{itemize}
\section{Related Work}

\subsection{Single Image Rain Streak Removal}\label{sec:Single image Deraining}
The main goal of single image rain streak removal is to reconstruct a clear image from a rainy image, further serving VBMS~\cite{ShirmohammadiF14, zhiqin}.
On account of the ill-posed nature of single image rain streak removal, traditional methods \cite{15,16} aim to leverage diverse image priors to impose additional constraints.
Nevertheless, the priors in these manually designed approaches are frequently derived from empirical observations, thus lacking the ability to comprehensively capture the inherent characteristics of clear images.
To address these challenges, many frameworks based on CNNs~\cite{argan,8} have emerged for single image rain streak removal and these frameworks have demonstrated satisfactory restoration results. 
To accurately capture the distribution of rainfall, recent studies have started to explore rainfall-related characteristics like direction \cite{8} and density \cite{9}. They optimize the network structure through the introduction of diverse computations \cite{10,8} and transfer mechanisms \cite{15,16} aimed at improving the removal of rain streak.
While CNNs~\cite{argan,8} perform better than manually designed prior-based methods, they face challenges in capturing long-distance dependencies because of the limitations of convolutional operations.
%
%
Our method separats rain streak provides a novel mechanism to enhance image restoration quality while addressing computational limitations of CNN-based~\cite{argan} and Transformer-based~\cite{ustn,mvksr} methods.

Addressing rain streak degradation is crucial for improving the VBMS in real-world applications, including object detection~\cite{mubiaojiance1} and segmentation~\cite{fenge}, which highlight the importance of image restoration as a critical preprocessing step. Our work aligns with these advancements by focusing on the unique problem of image rain streak removal and its broader applications for instrumentation and measurement systems~\cite{celiang}.

%

\subsection{Vision Transformer}

Given the Transformer’s remarkable success in natural language processing (NLP) \cite{17} and advanced visual tasks \cite{yinyong1,yinyong2,yinyong3,yinyong4,yinyong5}.
It has been adopted in image restoration.
Compared to previous CNNs, Transformers excel at capturing non-local information, thereby demonstrating superior performance in image restoration tasks.
Recently, Xiao et al. \cite{21} and their teams carefully develop a single image deraining network, which is a dual Transformer-based method on windows and space. 
Most current deep learning methods focus on self-concerned mechanisms of dense dot products.
However, this approach has a drawback, unimportant or extraneous features with low weights can compromise the accuracy of the attention map.
The current attention mechanism is global, primarily aimed at capturing long-range dependencies across upper and lower layers during modeling.
It does not effectively model local features, which may result in output features that still exhibit rain streak.
Attention is widely used in various tasks, including segmentation \cite{fenge} and detection \cite{mubiaojiance1}.
The self-attention mechanism first gained significant attention in natural language processing \cite{25}.
Self-attention-based method \cite{27} determines the context encoding of a position by calculating a weighted aggregation of embeddings.
Zhao et al.\cite{28} propose PSANet, which aggregates contextual information for each position through predicted attention maps.
%
%
%
Fu et al.\cite{30} propose DANet, which applies dual attention mechanism  to collect  useful information from feature maps, requiring more computational resources than other methods.
%
%
Wang et al. \cite{31} is the first to apply the self-attention mechanism to the computer vision (CV) tasks and introduces the non-local method.
However, non-local blocks require computing correlations between all pixels, resulting in high time and spatial complexity, increased computing resource consumption, and excessive redundant information.
Therefore, exploring algorithms that optimize network structures to effectively remove redundant information is crucial.

\subsection{Expectation Maximization Algorithm}
The Expectation Maximization (EM) \cite{32} is an effective method to solve the implicit variable optimization problem.
Consider ${X={x_{1},x_{2},...,x_{n}}}$ as a dataset with $N$ observation samples, where each $x_{i}$ is associated with the underlying variable $z_{i}$.
The EM algorithm is structured around two main steps: the Expectation step and the Maximization step. 
Its main goal is to estimate the model parameters through this process.
In step Expectation, We use the current parameters $\eta^{\text{former}}$ to compute $p(X,Z \mid \eta)$. The posterior distribution of $Z$ is then derived based on this. 
We then use the subsequent data to evaluate the probability of the dataset $L(\eta\mid\eta^{{former}})$, given by the following equation:
\begin{equation}
\small 
L\left(\eta,\eta^{former}\right)=Expect\left[\ln p(X,Z\mid\eta)\mid X,\eta^{former}\right],\end{equation}
Then in step Maximization, the modified parameters are determined by maximizing the function $\eta^{{current}}$:
\begin{equation}\eta^{{current}}=\arg\max_\eta L\left(\eta,\eta^{former}\right),\end{equation}
The Expectation step calculates the expected value of $z_{nq}$, representing the contribution value of basis ${\mu}$ to $x_{n}$.
Given the $\mu_{q}$, the likelihood of $x_{n}$ can be expressed$:$
\begin{equation}p({x}_{n}\big|\mu_{q})=\Phi({x}_{n},\mu_q),\end{equation}
wherein $\Phi$ denotes standard kernel function. 

In experiments, the Expectation step is implemented using matrix multiplication followed by a softmax layer.
The operation of the Expectation step in $t$ iteration can be expressed as:
\begin{equation}Z^{({t})}={softmax}(\beta{X}(\mu^{({t}-1)})^{{T}}),\label{con:em1}\end{equation}
in which $softmax(\cdot)$ means the softmax function; $\beta$ is a parameter that regulates $Z$.

Using the calculated $Z$, the M step updates ${\mu}$.
To make sure that $X$ and bases are in the same space, the bases ${\mu}$ are undated by performing a weighted sum of $X$. $\mu_{q}$ is updated as $\text{:}$
\begin{equation}{\mu}_q^{(t)}=\frac{z_{nq}^{(t)}{x}_n}{\sum_{m=1}^Nz_{mq}^{(t)}},\label{con:em2}\end{equation}

Li et al. \cite{32.5} reconfigure the self-attention mechanism through an expectation maximization algorithm approach and present an expectation maximization attention mechanism tailored for semantic segmentation.
The newly proposed Expectation Maximization Attention mechanism (EMA) module shows robustness to input variations and is efficient in terms of memory usage and computational cost.

\begin{figure}[!t] 
\includegraphics[width=\linewidth]{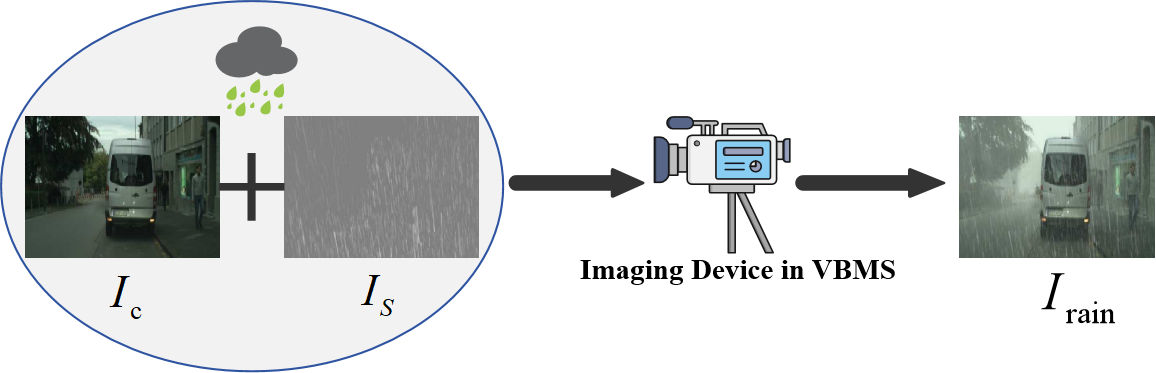} 
\vspace{-6mm}
\caption{Illustration of the rain imaging degraded model in VBMS.}
\label{fig:pYU}
\vspace{-4mm}
\end{figure}

\begin{figure*}[!t] 
\includegraphics[width=0.98\linewidth]{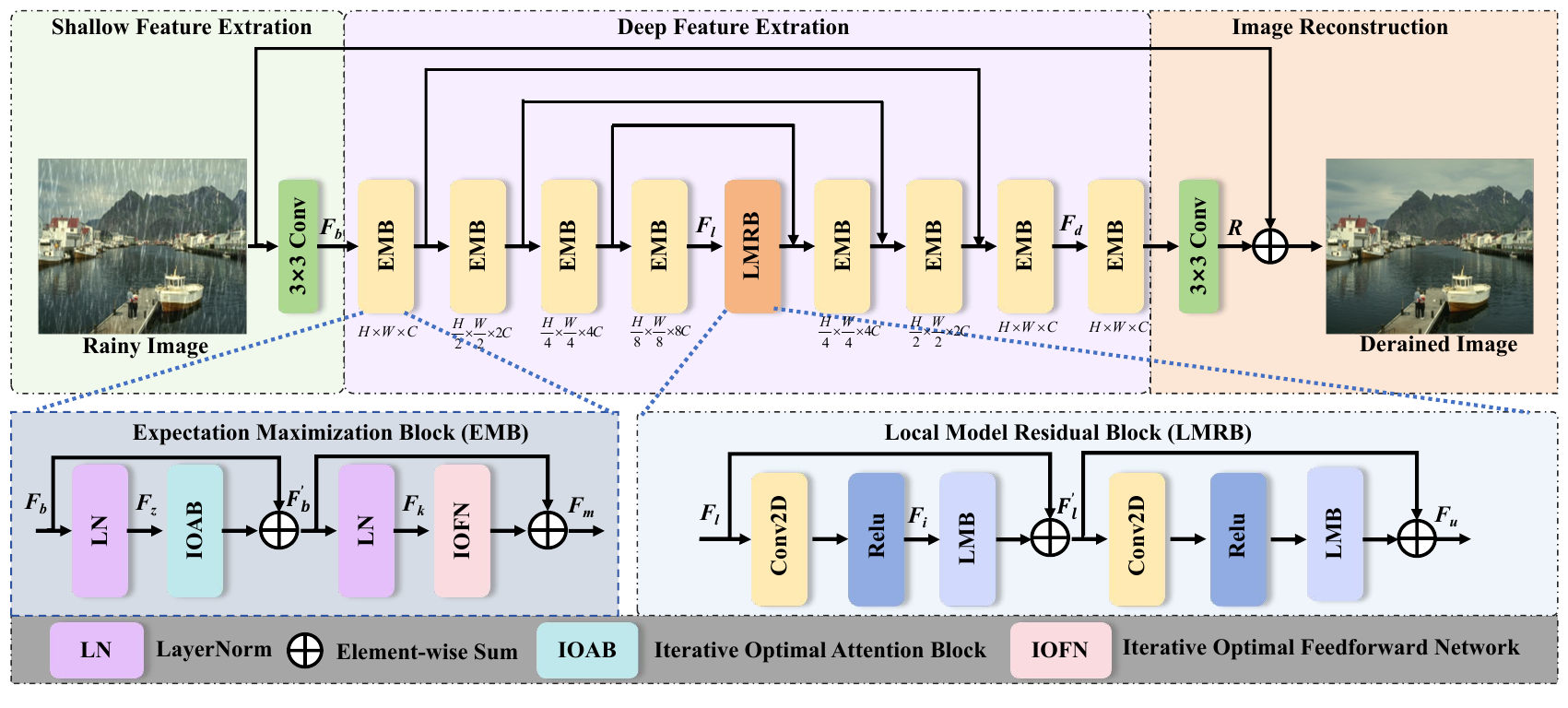}
\vspace{-6mm}
\caption{The overall architecture of EMResformer. 
Our EMResformer consists of three major components: Shallow Feature Extraction Module, Deep Feature Extraction Module, and Image Reconstruction Module.
Given a rainy image input, a 3$\times$3 convolution layer is employed to extract the shallow features.
Then, we use a series of Expectation Maximization Blocks as the Deep Feature Extraction Module to extract deeper features. 
Additionally, we use the Local Model Residual Block to enhance the network's capability to model local features.
Finally, in the image reconstruction stage, we use a 3$\times$3 convolution and add the original input to generate a clean background image.
}
\label{fig:p2}
\vspace{-2mm}
\end{figure*}
Drawing inspiration from EMA, we design the EMResformer that can more effectively eliminate redundant information and enhance the local model capability of the network, guiding the rain streak removal network to learn more useful image details and achieve cleaner deraining results.
%
%

%
%
\section{Proposed Approach}

We aim to effectively separate rain streak from backgrounds in a given rainy image.
To this end, we propose an Expectation Maximization Reconstruction Transformer (EMResformer) for single image deraining as described in Section~\ref{sec: overall}.
%
To effectively eliminate redundant information and generate attention maps with a concentrated distribution, we propose an Expectation Maximization Block (EMB) into the backbone network, as described in Section~\ref{sec: EM Block}.
%
To improve the focus on local information, we propose a Local Model Residual Block (LMRB) for providing the local model capability of the network, as detailed in Section~\ref{sec:Local Model Residual Block}.
\vspace{-3mm}
\subsection{Physical Imaging Model for Rainy Conditions}\label{sec: Physical Imaging Model for Rainy Conditions}

To characterize the impact of rain on image quality in VBMS, we present a physical imaging model that simulates the degradation caused by rain streaks and other weather-related effects.
As shown in Fig.~\ref{fig:pYU}, the model incorporates the clean image $I_{c}$
and and the rain streak  $I_{s}$
resulting in the rainy image  $I_{rain}$
\begin{equation}I_{rain}=I_{c}+I_{s}.\end{equation}
This model captures the main degradation caused by rain, leading to reduced visibility and measurement accuracy in VBMS applications.

\vspace{-2mm}
\subsection{Overall Pipeline}\label{sec: overall}
The proposed network aims to learn a non-linear $f$,
which directly describes the mapping relationship between the rainy image and the rain streak. 
Fig.~\ref{fig:p2} shows the overall pipeline of the proposed EMResformer, which mainly consists of three components: Shallow Feature Extraction Module, Deep Feature Extraction Module and Image Reconstruction Module.
Given a rainy image $I_{rain}\in\mathbb{R}^{H\times W\times3}$, where $H$ and $W$ represent the spatial resolution of the feature map. 
We first use a 3$\times$3 convolution to extract shallow features $F_{b}\in\mathbb{R}^{H\times W\times C}$.

Subsequently, deep feature extraction is executed within a four-layer symmetric encoder-decoder architecture.
During the encoding phase, the spatial dimensions are progressively diminished, concurrent with an increase in channel capacity.
In particular, the input shallow features $ F_{b} \in \mathbb{R}^{H \times W \times C} $, characterized by high resolution, are processed through an encoder comprising four EMB layers to derive low-resolution latent features $ F_{l} \in \mathbb{R}^{\frac{H}{8} \times \frac{W}{8} \times 8C} $. 
Concurrently, the LMRB is utilized to augment the local modeling capabilities effectively.
In the decoding phase, the low resolution features $ F_{l} \in \mathbb{R}^{\frac{H}{8} \times \frac{W}{8} \times 8C} $ serve the purpose of inputs to a decoder composed of four layers of EMB architecture. This decoder progressively recovers high-resolution representations to yield the deeper features $F_{{d}}\in\mathbb{R}^{H\times W\times C}$.

To aid the recovery process, encoder is linked to decoder via skip connections, ensuring stable training.
This aids in maintaining fine structural and texture.
In the next step, further enrich the deeper features $F_{d}$ during the refinement stage of operations with high spatial resolution.

Finally, in the image reconstruction stage, a convolution layer is used to transform the refined features into a residual image.
The residual image $R\in\mathbb{R}^{H\times W\times3}$ is then added to the degraded image to obtain the restored image.
%
%
The final reconstruction result is obtained by the following formula$:$
\begin{equation}I_{{derain}}=\mathcal{G}(I_{rain})+I_{rain},\end{equation}
where $\mathcal{G(\cdot)}$ constitutes the entire network. 
The loss function is defined as follows:
\begin{equation}{L}=\|I_{derain}-I_{gt}\|,\end{equation}
where $I_{{gt}}$ stands for clean, rainless image, and $\|{\cdot}\|$ represents the loss of SSIM.

%
%
\subsection{Expectation Maximization Block}\label{sec: EM Block}
%
%
The standard Transformer \cite{35}, which computes self-attention globally across all tokens, is not well-suited for image recovery due to its inclusion of excessive redundant information. 
To address this limitation, we introduce an expectation maximization block as a feature extraction unit, leveraging the EM algorithm to iteratively refine the network's attention distribution.
\begin{figure*}[!t]
\centering
\includegraphics[width=0.98\linewidth]{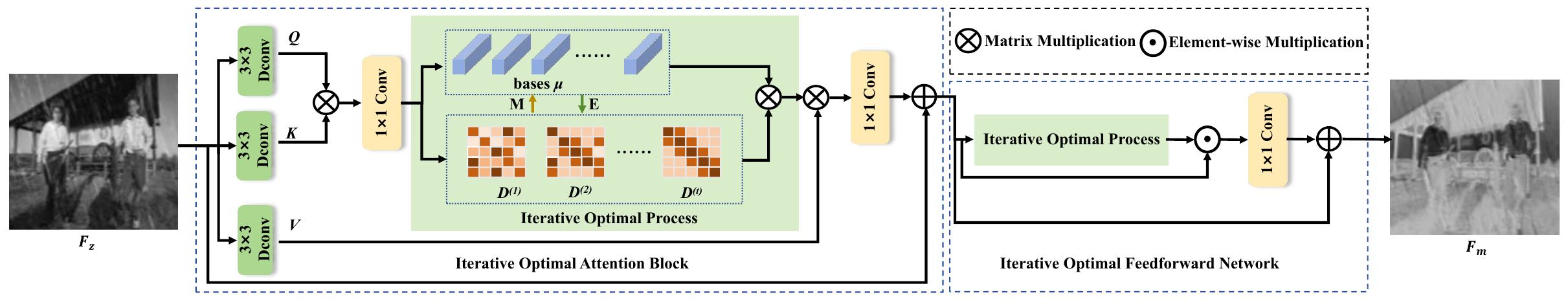}
\vspace{-4mm}
\caption{The architecture of the Expectation Maximization Block (EMB). EMB is used to effectively extract features from images, facilitating the clear background images.
}
\label{fig:p3}
\vspace{-2mm}
\end{figure*}
%

%
%
Within each EMB, in contrast to the standard self-attention \cite{35}, we design an Iterative Optimal Attention Block (IOAB) and an Iterative Optimal Feedforward Network (IOFN) to implement attention and feedforward networks, with the goal of enhancing feature aggregation processes. The overall workflow of EMB is as follows:
\begin{equation}F_{b}^{'}=F_{b}+f_{IOAB}(LN(F_{b})),\label{con:inventoryflow}\end{equation}
\begin{equation}F_{m}=F_{b}^{'}+f_{IOFN}(LN(F_{b}^{'})),\label{con:inventoryflow1}\end{equation}
where $F_{b}$ denotes the input of the first EMB; $F_{b}^{'}$ signifies the intermediate features of the EMB; $F_{m}$ represents the output of EMB; $LN(\cdot)$ indicates layer normalization; $f_{IOAB}(\cdot)$ denotes the Iterative Optimal Attention Block; and $f_{IOFN}(\cdot)$ signifies the Iterative Optimal Feedforward Network.

\textbf{Iterative Optimal Attention Block.} 
%
Traditional attention mechanisms frequently produce feature maps with significant redundancy, which hinders the effective restoration of clean background images.
%
To tackle this challenge, we introduce an adaptive weight adjustment mechanism that dynamically modifies feature importance throughout the iterative optimization process of the expectation maximization (EM) algorithm.
This method efficiently removes redundant features.
%
The Iterative Optimal Attention Block is depicted in Figure~\ref{fig:p3}.
We have revisited the self-attention mechanism.
%
%

Traditional attention mechanisms mainly utilize query $Q$, keyword $K$, and value $V$.
The traditional attention mechanism formula can be expressed:
\begin{equation}A{ttention}(Q,K,V)=softmax(\frac{QK^{T}}{\lambda})V,\end{equation}
$\lambda$ is computed as $\lambda = \sqrt{d}$, with an optional temperature factor denoted by $D$.
%
Multiple heads process the new $Q$, $K$, and $V$ to generate outputs within the ${d}=C/K$ channel dimension. 
These outputs are concatenated and linearly transformed to produce the final result for the all heads.
%
It is noteworthy that the prevalent self-attention paradigm is grounded in dense full connectivity, necessitating the computation of attention graphs for every query-key pair.
%
In our network, we introduce IOAB as a substitute for the conventional self-attention mechanism, thereby reducing the incorporation of irrelevant and redundant information in feature interactions.
%

%
We first encode the channel context with a 3$\times$3 deep convolution.
%
Drawing inspiration from \cite{36}, we use self-attention mechanisms across channels rather than spatial dimensions to reduce computational complexity in both time and memory.
%
Subsequently, we compute the attention score matrix by multiplying the generated query $Q$ and key $K$. This matrix is then combined with a compact set of bases $\mu$ to execute the maximum expected algorithm, updating the set of bases in Maximization step. 
In Expectation step, we update the attention matrix by multiplying $Q$ and $K$, reinforcing the focused regions of the attention matrix, resulting in a matrix with reduced redundancy. This leads to the generation of a feature attention map.
%
Matrix multiplication with the value $V$ follows, and the process concludes with residual concatenation via a 1$\times$1 convolution, which serves as the final output.
%

%
The IOAB alternates between the E step and M step over $t$ iterations.
Drawing on formula (\ref{con:em1}) and (\ref{con:em2}), the updated $QK^{T}$ is expressed as follows:
\begin{equation}D^{(t)}=(QK^{T})^{t}=Z^{(t)}\mu^{(t)},\end{equation}
where $D^{(t)}$ represents the attention weight distribution of optimal iterative attention; $t$ denotes the number of iterations for the EM operation.

The $A{ttention}(Q,K,V)$ process can be described as$\text{:}$
\begin{equation}A{ttention}(Q,K,V)=(\frac{D^{(t)}}{\lambda})V,\end{equation}

Drawing on formula (\ref{con:inventoryflow}), giving an input tensor $F_{z}$, the process of IOAB is defined as$\text{:}$
\begin{equation}F_{z}=LN(F_{b}),\end{equation}
\begin{equation}F_{z}^{'}=F_{z}+(\frac{D^{(t)}}{\lambda})V,\end{equation}
where $F_{z}^{'}$ denotes the output features of the IOAB.
%

%
\textbf{Iterative Optimal Feedforward Network.} 
%
In the hidden layers of feedforward neural networks, significant fuzzy computation is often observed, leading to redundant information processing.  
%
To iteratively optimize the parameters and structure of the hidden layers, we introduce an iterative optimization feedforward network that employs the Expectation Maximization (EM) algorithm, as illustrated in Fig.~\ref{fig:p3}.
This approach dynamically adjusts the network's representational capacity and complexity, thereby enhancing the model's effectiveness in handling complex data and learning underlying distributions.
%

Feedforward network (FFN) \cite{37}, operates on each pixel position separately and equally to transform features. 
They employ two branches: one to broaden the feature channels and another to condense them back to the original input dimensions.
%
%
We propose two key modifications to the FFN to enhance the learning of representations: the maximum expectation mechanism and deep convolution.
Similar to IOAB, we also employ deep convolution in IOFN to encode information. 
%
By decoupling $G$ from the input features, the maximum expectation algorithm is executed with bases $\gamma$, and the optimal solution $G$ with reduced redundancy is progressively obtained.
%
Subsequently, pointwise multiplication is conducted using divided input feature, followed by a connection achieved through a 1$\times$1 convolution, which outputs the input. 
Drawing on formula (\ref{con:em1}) and (\ref{con:em2}), after the EM algorithm, $G$ can be described as$\text{:}$
\begin{equation}G^{(t)}=(W_{d}^{1}W_{p}^{1}(LN(X)))^{t}=Z^{(t)}\gamma^{(t)},\end{equation}
in which $G^{(t)}$ represents the result of the iteration optimal process.

Drawing on formula (\ref{con:inventoryflow1}), giving an input tensor $F_{k}$, the process of IOFN is defined as $\text{:}$
\begin{equation}F_{k}=LN(F_{b}^{'}),\end{equation}
\begin{equation}F_{k}^{'}=G^{(t)}\odot W_{d}^{2}W_{p}^{2}(F_{k})+F_{k},\end{equation}
where $\odot$ represents elementwise multiplication; $F_{k}^{'}$ denotes the results produced by the IOFN.

\begin{figure}[!t] 
\centering
\includegraphics[width=\linewidth]{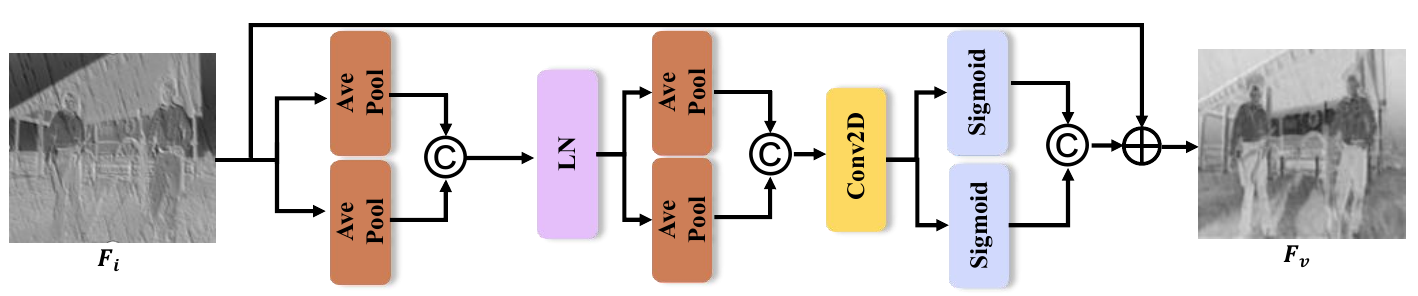}
\vspace{-6mm}
\caption{The structure of the Local Model Block (LMB). 
LMB comprises a sequence of pooling, activation, and convolution operations to improve the network's local model capability for recovering clean background images.}
\label{fig:p4}
\vspace{-2mm}
\end{figure}
\begin{figure*}[!t]
\centering
\footnotesize 
\renewcommand{\arraystretch}{1.0} 
\setlength{\tabcolsep}{1pt}       
\begin{tabular}{ccccccc}
\includegraphics[width=0.14\linewidth]{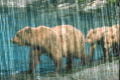} &
\includegraphics[width=0.14\linewidth]{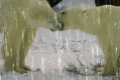} &
\includegraphics[width=0.14\linewidth]{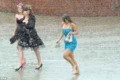} &
\includegraphics[width=0.14\linewidth]{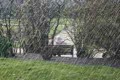} 
&\includegraphics[width=0.14\linewidth]{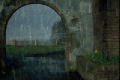} 
&\includegraphics[width=0.14\linewidth]{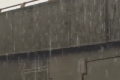} 
&\includegraphics[width=0.14\linewidth]{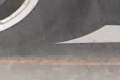} \\
(a) Rain200H  & (b) Rain200L  & (c) Rain1200 & (d) Rain1400 & (e) Rain12& (f) MPID& (g) SPA-Data \\

\end{tabular}
\vspace{-3mm}
\caption{Five synthesized datasets containing rain streak and two real-world dataset. These exhibit a wide range of rain streak
orientations, sizes, and shapes.}
\label{fig:p91}
\vspace{-4mm}
\end{figure*}

\vspace{-2mm}
\subsection{Local Model Residual Block}\label{sec:Local Model Residual Block}
%
Current network architectures commonly employ global attention mechanisms to capture long-range dependencies, yet they may overlook the critical importance of local information in reconstructing fine texture details.
%
For EMResformer, Iterative Optimal Attention serves as a global attention mechanism. 
%
During the modeling process, it emphasizes long-distance dependencies between upper and lower layers but fails to effectively model local contexts.
To mitigate this issue, we introduce a Local Model Residual Block, as depicted in Figure~\ref{fig:p2}.
%
By incorporating positional information into channel attention, we can use the network effectively to target larger areas while minimizing computational costs.
Given an input tensor $F_{l}$, LMRB can be expressed as$\text{:}$
\begin{equation}F_{l}^{'}=F_{l}+f_{LMB}(Relu(Conv(F_{l}))),\label{con:inventoryflow2}\end{equation}
\begin{equation}F_{u}=F_{l}^{'}+f_{LMB}(Relu(Conv(F_{l}^{'})))),\end{equation}
%
where $F_{l}^{'}$ denotes the intermediate features of the LMRB; $F_{u}$ represents the output of LMRB; the $Conv(\cdot)$ denotes the 1×1 convolution function; the $Relu(\cdot)$ denotes the ReLU activation function; and $f_{LMB}(\cdot)$ signifies the Local Model Block.

\textbf{Local Model Block.}
The Local Model Block (LMB) is a pivotal component of the LMRB, as illustrated in Figure~\ref{fig:p4}.
%
To address the loss caused by 2D global pooling, We split channel attention into two 1D encoding streams to preserve spatial coordinate information within the attention map.
%
One-dimensional global pooling is applied in both vertical and horizontal directions to create two distinct feature maps, each capturing different directional perceptions of the input.
%
These feature maps are then used to generate two attention maps, which capture long-range dependencies in different spatial directions.
%
Consequently, positional information is retained in the attention maps, which are then used to enhance the input feature map's representational capacity through multiplication.

Drawing on formula (\ref{con:inventoryflow2}), giving an input tensor $F_{i}$, LMB can be expressed as$\text{:}$
\begin{equation}F_{i}=Relu(Conv(F_{l})),\end{equation}
\begin{equation}F_{i}^{'}=F_{i}+Sigmoid((AvePool(LN(AvePool(F_{i})))),\end{equation}
%
where $Sigmoid(\cdot)$ denotes the Sigmoid operation; $F_{i}^{'}$ denotes the output features from the LMB; AvePool stands for average pooling operation.

%
%
The LMB process can be conceptualized as a local attention mechanism, conferring the current network model with a shared global-local modeling capability. 
%
To further enhance local modeling capabilities, it is structured as a residual module, which ensures model stability during training and improves the robustness of the restored image.
%

%

%

%

%
%
\begin{table}[!t]
    \centering
    \caption{Detailed description of seven datasets and train--test information}
    \vspace{-2mm}
    \resizebox{\linewidth}{!}{ 
    \renewcommand{\arraystretch}{1.25}
    \begin{tabular}{@{}lcccccc@{}}
    \toprule
    Dataset & Image size & Image type & Data & Train Samples & Test Samples \\ \midrule
    Rain200H \cite{rain200}      & (481, 321)   & Synthetic  & Train + Test & 1800    & 200   \\
    Rain200L \cite{rain200}     & (481, 321) & Synthetic  & Train + Test & 1800    & 200   \\
    Rain1200 \cite{rain1200}  & (512, 512)   & Synthetic  & Train + Test & 12000  & 1200    \\
    Rain1400 \cite{7} & - & Synthetic  & Train + Test & 12600    & 1400     \\
    Rain12 \cite{rain12}    & (481, 321) & Synthetic  & Test & -   & 12    \\
    SPA-Data \cite{spadata}        & (512, 512)   & Real  & Test & -  & 1000    \\
    MPID \cite{mpid}       & -   & Real  & Test         & -  & 256   \\ \bottomrule
\end{tabular}
    }
    \label{tab:t414}
    \vspace{-4mm}
\end{table}

\begin{table*}[!t]
\centering
\caption{
Quantitative results on widely-used benchmarks. The \textbf{best} results are marked in bold.
Higher PSNR and SSIM values (Mean ± Std) reflect improved performance.
All the results are evaluated on the Y channel.}
\vspace{-2mm}
\renewcommand{\arraystretch}{1.25}
\resizebox{\textwidth}{!}{
\begin{tabular}{c|ccccc}
\hline
\multirow{2}{*}{Method}     & Rain200H~\cite{rain200}        & Rain200L~\cite{rain200}        & Rain1200~\cite{rain1200}       & Rain1400~\cite{7}     & Rain12~\cite{rain12}     
\\ \cline{2-6}
            & PSNR/SSIM     & PSNR/SSIM      & PSNR/SSIM     & PSNR/SSIM     & PSNR/SSIM   \\\hline
RESCAN \cite{11}           & 26.65±3.498/0.841±0.015 & 36.99±3.595/0.978±0.007  & 32.12±3.510/0.902±0.010 & 30.96±3.300/0.911±0.012 & 32.96±3.400/0.954±0.009 \\  
NLEND \cite{38.02}         & 27.32±3.960/0.890±0.014 & 36.48±3.840/0.979±0.006  & 32.47±3.650/0.919±0.011 & 31.01±3.530/0.920±0.010 & 33.02±3.700/0.961±0.008 \\  
SSIR \cite{38.03}          & 14.42±3.220/0.450±0.040 & 23.47±4.990/0.802±0.030  & 25.42±4.020/0.771±0.025 & 25.77±4.210/0.822±0.022 & 24.13±4.500/0.776±0.028 \\  
PreNet \cite{12}           & 27.52±2.590/0.866±0.012 & 34.26±2.800/0.966±0.008  & 30.45±2.510/0.870±0.014 & 30.98±2.430/0.915±0.011 & 35.09±2.700/0.940±0.009 \\  
SPANet \cite{38.05}        & 25.48±4.370/0.858±0.014 & 36.07±3.900/0.977±0.007  & 27.09±3.850/0.808±0.018 & 29.00±3.700/0.889±0.012 & 33.21±3.900/0.954±0.009 \\  
DCSFN \cite{38.06}         & 28.46±2.080/0.901±0.011 & 37.85±2.630/0.984±0.006  & 32.27±2.400/0.922±0.010 & 31.49±2.200/0.927±0.009 & 35.80±2.300/0.968±0.007 \\  
MSPFN \cite{10}            & 25.55±5.330/0.803±0.017 & 30.36±4.000/0.921±0.011  & 30.38±3.580/0.886±0.013 & 31.51±3.800/0.920±0.010 & 34.25±4.000/0.946±0.008 \\  
DRDNet \cite{38.08}        & 15.10±2.730/0.502±0.038 & 37.46±3.610/0.980±0.005  & 28.38±3.860/0.857±0.014 & 28.35±3.700/0.857±0.013 & 25.43±4.120/0.755±0.030 \\  
RCDNet \cite{38.09}        & 28.68±3.220/0.890±0.013 & 38.40±2.930/0.984±0.005  & 32.27±2.800/0.911±0.011 & 31.01±2.700/0.916±0.010 & 31.03±2.900/0.906±0.012 \\  
Syn2Real \cite{15}         & 14.49±3.720/0.402±0.045 & 31.03±3.900/0.936±0.012  & 28.81±3.900/0.840±0.018 & 28.58±3.870/0.858±0.015 & 31.03±4.000/0.906±0.011 \\  
MPRNet \cite{38.11}        & 29.94±3.230/0.915±0.010 & 36.60±3.250/0.978±0.006  & 33.65±3.080/0.931±0.009 & 32.25±2.960/0.932±0.008 & 36.57±3.150/0.956±0.007 \\  
OHCNet \cite{38.12}        & 29.98±3.180/0.921±0.009 & 39.28±2.850/0.987±0.005  & 33.71±2.950/0.932±0.008 & 32.61±2.830/0.933±0.007 & 36.85±3.010/0.971±0.006 \\  
RLP \cite{38.15}          & 28.35±3.820/0.873±0.012 & 34.64±3.120/0.956±0.008  & 31.67±3.200/0.904±0.010 & 30.63±3.110/0.903±0.011 & 34.32±3.300/0.954±0.009 \\  
Restormer \cite{36}        & 29.98±2.800/0.920±0.009 & 38.23±2.800/0.972±0.007  & 33.02±2.900/0.923±0.009 & 31.23±2.800/0.932±0.008 & 35.62±2.900/0.958±0.007 \\  
MFDNet \cite{38.14}        & 29.28±3.230/0.914±0.010 & 38.58±2.920/0.985±0.005  & 33.01±2.900/0.925±0.008 & 31.91±2.880/0.926±0.009 & 36.15±3.100/0.964±0.007 \\  
MSHFN \cite{38.13}         & 26.06±3.480/0.853±0.015 & 31.35±3.650/0.946±0.011  & 32.46±3.500/0.919±0.010 & 32.32±3.450/0.931±0.009 & 35.35±3.700/0.956±0.007 \\  
MFFDNet \cite{9}              & \textbf{30.72±2.852}/0.908±0.013 & 39.44±2.88/0.987±0.006  & -/- & 31.85±3.02/0.935±0.010 & -/- \\   
\textbf{Ours}             & 30.30±2.950/\textbf{0.924±0.008} & \textbf{39.60±2.850}/\textbf{0.990±0.004}
& \textbf{34.03±3.100}/\textbf{0.935±0.007} & \textbf{32.93±2.950}/\textbf{0.936±0.006} & \textbf{37.17±3.150}/\textbf{0.974±0.005} \\   
\hline
\end{tabular}
}
\label{tab:t1}
 \vspace{-4mm}
\end{table*}

\section{Experiments and Analysis}
In this section, we perform a series of experiments to evaluate the effectiveness of the proposed EMResformer.
\vspace{-2mm}
\subsection{Dataset and Evaluation Metrics}
\textbf{Dataset.} We evaluate the model on standard benchmarks, including the synthetic datasets Rain200L~\cite{rain200}, Rain200H~\cite{rain200}, Rain1200~\cite{rain1200}, Rain1400~\cite{7}, Rain12~\cite{rain12}, and the real datasets SPA-Data~\cite{spadata} and MPID~\cite{mpid}. Table~\ref{tab:t414} provides a detailed description of each dataset and the train-test information. 
We also provide a comprehensive overview of these datasets and present visual demonstrations from different datasets in Fig.~\ref{fig:p91}, which feature various rain streak directions, sizes, and shapes.
To further evaluate the performance of our EMResformer on classical high-level vision tasks (object detection and semantic segmentation), we additionally include COCO350~\cite{COCO} and BDD350~\cite{BDD} as test datasets.
\textbf{Evaluation Metrics.} We utilize reference-based evaluation metrics (SSIM and PSNR) and nonreference metrics (NIQE and BRISQUE), as these metrics are widely used in the field of visual measurement~\cite{0.2,0.3,0.4} to evaluate image quality and structural similarity. 
We also use the additional metrics commonly used in vision-based measurement applications, including Precision, Recall, mAP@0.5, and mAP@0.5:0.95.

\textit{1)} Reference-Based Metrics: We utilize PSNR and SSIM to evaluate the performance of our method. The details of these reference-based metrics are described below.

\textcolor{red}{(a)} Peak Signal-to-Noise Ratio (PSNR)~\cite{PSNR_thu} is a commonly used metric to assess the similarity between corresponding pixels of the derained image $I_{out}$ and the ground truth
image $I_{gt}$. MSE$(I_{out}, I_{gt})$ denotes the mean square error between the
derained image and the ground truth. A higher PSNR value indicates better derained image quality,
as expressed mathematically in the
following:
\begin{equation}\mathrm{PSNR}(I_{out},I_{gt})=10\log_{10}\biggl(\frac{{255}^2}{\mathrm{MSE}(I_{out},I_{gt})}\biggr), \end{equation}

\textcolor{red}{(b)} The Structural Similarity Index (SSIM)~\cite{psnr} evaluates the similarity between the derained image $I_{out}$ and the ground truth
image $I_{gt}$ based on three components: luminance, contrast, and structure. Its value ranges from 0 to 1, where a higher SSIM value corresponds to better derained image quality. This mathematical expression is represented by the
following:
\begin{equation}
\hspace{-2mm}
\small  
\mathrm{SSIM}(I_{out},\tau) = 
\frac{(2\mu_{I_{out}}\mu_{I_{gt}}+\delta_1)(2\sigma_{I_{out}I_{gt}}+\delta_2)}%
{\left(\mu_{I_{out}}^2+\mu_{I_{gt}}^2+\delta_1\right)\left(\sigma_{I_{out}}^2+\sigma_{I_{gt}}^2+\delta_2\right)},
\end{equation}
where $\mu_{I_{out}}$, $\mu_{I_{gt}}$, $\sigma_{I_{out}}$, and $\sigma_{I_{gt}}$ denote the mean and
standard deviation of derained image and ground truth
image. $\delta_1$ and $\delta_2$ are constants used to prevent division by zero.

\textit{2)} Nonreference Metrics: We utilize NIQE and BRISQUE to evaluate the performance of our method. The details of these nonreference metrics are described below.

(a) The Natural Image Quality Evaluator (NIQE)~\cite{niqe} assesses image quality by measuring deviations from the statistical characteristics typically found in natural images. A lower NIQE score indicates higher image quality.

(b) The Blind/Referenceless Image Spatial Quality Evaluator (BRISQUE)~\cite{brisque} evaluates image quality by modeling image coefficients with a Gaussian distribution and using a support vector machine (SVM) for assessment. A lower BRISQUE score indicates better image quality.

\textit{3)} Vision-Based Measurement Metrics: We use precision, Recall, mAP@0.5, and mAP@0.5:0.95 to evaluate the performance of object detection. The details of these nonreference metrics are described below.

(a) Precision and Recall are standard metrics for evaluating object detection tasks. Precision measures the accuracy of the positive predictions, while Recall measures the ability of the model to detect all relevant instances.

(b) mAP@0.5 calculates the average precision for object detection at a threshold of 0.5, giving an overall measure of the model's detection accuracy.

(c) mAP@0.5:0.95 extends mAP by averaging the precision across multiple IoU thresholds from 0.5 to 0.95, providing a more stringent and comprehensive evaluation of detection performance.

\subsection{Comparison Methods} We benchmark EMResformer against several state-of-the-art methods including RESCAN \cite{11}, NLEDN \cite{38.02}, SSIR \cite{38.03}, PReNet \cite{12}, SPANet \cite{38.05}, DCSFN \cite{38.06}, MSPFN \cite{10}, DRDNet \cite{38.08}, RCDNet \cite{38.09}, Syn2Real \cite{15}, MPRNet \cite{38.11}, OHCNet \cite{38.12}, RPL \cite{38.15}, Restormer \cite{36}, MSHFN \cite{38.13}, MFDNet \cite{38.14} and MFFDNet \cite{9}. We evaluate their pre-trained models in our comparisons.
%
%

%
\begin{figure*}[!t]
\begin{center}
\begin{tabular}{ccccccc}
\hspace{-1.5mm}\includegraphics[width = 0.14\linewidth]{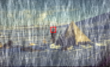} &\hspace{-4.5mm}
\includegraphics[width = 0.14\linewidth]{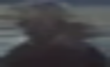}
&\hspace{-4.5mm}
\includegraphics[width = 0.14\linewidth]{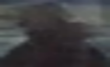} &\hspace{-4.5mm}
\includegraphics[width = 0.14\linewidth]{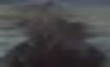} &\hspace{-4.5mm}
\includegraphics[width = 0.14\linewidth]{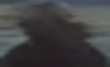} &\hspace{-4.5mm}
\includegraphics[width = 0.14\linewidth]{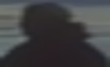} &\hspace{-4.5mm}
\includegraphics[width = 0.14\linewidth]{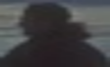} 
\\
\hspace{-1.5mm}\includegraphics[width = 0.14\linewidth]{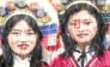} &\hspace{-4.5mm}
\includegraphics[width = 0.14\linewidth]{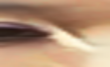} &\hspace{-4.5mm}
\includegraphics[width = 0.14\linewidth]{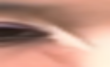} &\hspace{-4.5mm}
\includegraphics[width = 0.14\linewidth]{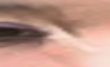} &\hspace{-4.5mm}
\includegraphics[width = 0.14\linewidth]{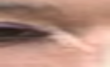} &\hspace{-4.5mm}
\includegraphics[width = 0.14\linewidth]{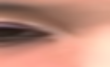} &\hspace{-4.5mm}
\includegraphics[width = 0.14\linewidth]{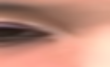}
\\
\hspace{-1.5mm}\includegraphics[width = 0.14\linewidth]{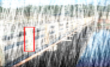} &\hspace{-4.5mm}
\includegraphics[width = 0.14\linewidth]{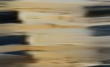} &\hspace{-4.5mm}
\includegraphics[width = 0.14\linewidth]{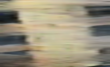} &\hspace{-4.5mm}
\includegraphics[width = 0.14\linewidth]{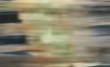} &\hspace{-4.5mm}
\includegraphics[width = 0.14\linewidth]{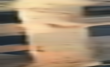} &\hspace{-4.5mm}
\includegraphics[width = 0.14\linewidth]{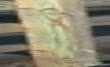} &\hspace{-4.5mm}
\includegraphics[width = 0.14\linewidth]{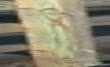}
\\
\hspace{-1.5mm}(a) Rain Image   &\hspace{-4.5mm} (b) SPANet &\hspace{-4.5mm}  (c) Restormer &\hspace{-4.5mm} (d) MFDNet &\hspace{-4.5mm} (e) MSHFN &\hspace{-4.5mm} (f) \textbf{Ours} &\hspace{-4.5mm} (g) GT
\end{tabular}
\end{center}
\vspace{-3mm}
\caption{Comparisons with SOTA methods on the synthetic datasets show that our EMResformer produces significantly clearer results with enhanced structural details.
}
\label{fig:p5}
\vspace{-3mm}
\end{figure*}
\begin{figure*}[!t]
\begin{center}
\begin{tabular}{ccccccc}
\hspace{-1.5mm}\includegraphics[width = 0.14\linewidth]{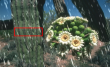} &\hspace{-4.5mm}
\includegraphics[width = 0.14\linewidth]{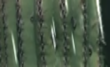}
&\hspace{-4.5mm}
\includegraphics[width = 0.14\linewidth]{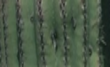} &\hspace{-4.5mm}
\includegraphics[width = 0.14\linewidth]{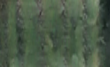} &\hspace{-4.5mm}
\includegraphics[width = 0.14\linewidth]{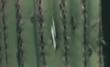} &\hspace{-4.5mm}
\includegraphics[width = 0.14\linewidth]{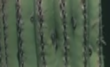} &\hspace{-4.5mm}
\includegraphics[width = 0.14\linewidth]{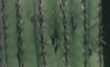} 
\\
\hspace{-1.5mm}\includegraphics[width = 0.14\linewidth]{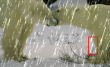} &\hspace{-4.5mm}
\includegraphics[width = 0.14\linewidth]{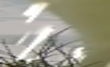} &\hspace{-4.5mm}
\includegraphics[width = 0.14\linewidth]{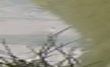} &\hspace{-4.5mm}
\includegraphics[width = 0.14\linewidth]{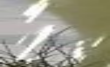} &\hspace{-4.5mm}
\includegraphics[width = 0.14\linewidth]{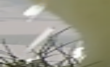} &\hspace{-4.5mm}
\includegraphics[width = 0.14\linewidth]{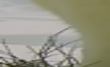} &\hspace{-4.5mm}
\includegraphics[width = 0.14\linewidth]{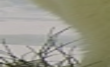}
\\
\hspace{-1.5mm}\includegraphics[width = 0.14\linewidth]{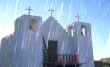} &\hspace{-4.5mm}
\includegraphics[width = 0.14\linewidth]{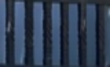} &\hspace{-4.5mm}
\includegraphics[width = 0.14\linewidth]{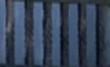} &\hspace{-4.5mm}
\includegraphics[width = 0.14\linewidth]{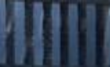} &\hspace{-4.5mm}
\includegraphics[width = 0.14\linewidth]{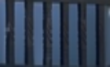} &\hspace{-4.5mm}
\includegraphics[width = 0.14\linewidth]{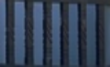} &\hspace{-4.5mm}
\includegraphics[width = 0.14\linewidth]{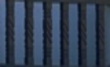}
\\
\hspace{-1.5mm}\includegraphics[width = 0.14\linewidth]{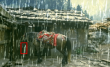} &\hspace{-4.5mm}
\includegraphics[width = 0.14\linewidth]{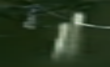} &\hspace{-4.5mm}
\includegraphics[width = 0.14\linewidth]{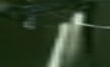} &\hspace{-4.5mm}
\includegraphics[width = 0.14\linewidth]{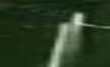} &\hspace{-4.5mm}
\includegraphics[width = 0.14\linewidth]{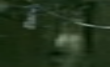} &\hspace{-4.5mm}
\includegraphics[width = 0.14\linewidth]{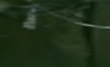} &\hspace{-4.5mm}
\includegraphics[width = 0.14\linewidth]{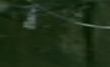}
\\
\hspace{-1.5mm}(a) Rain Image   &\hspace{-4.5mm} (b) SPANet &\hspace{-4.5mm}  (c) Restormer &\hspace{-4.5mm} (d) MFDNet&\hspace{-4.5mm} (e) MSHFN &\hspace{-4.5mm} (f) \textbf{Ours} &\hspace{-4.5mm} (g) GT
\end{tabular}
\end{center}
\vspace{-3mm}
\caption{Comparisons with the SOTA method on synthetic datasets show that our EMResformer is better at clearing rain streak.}
\label{fig:p6}
 \vspace{-4mm}
\end{figure*}

\subsection{Implementation Details}
%
In our model, the EMResformer is a four-level symmetric encoding and decoding framework. Progressing from level one to level four, the number of Encoder-Decoder Blocks (EMB) $\{N1, N2, N3, N4\}$ is set to $\{9, 6, 3, 0\}$. 
During the training process, we employ the AdamW optimizer with a patch size of $128$ for each iteration. 
The learning rate starts at $0.0001$, with the training spanning 500 epochs.
We set bach size to 4.
We implement the network using PyTorch and train it on the NVIDIA 3090Ti.
We chose this hardware and framework for their computational efficiency and compatibility with existing code bases.
Expanding upon these details provides a deeper understanding of the model architecture, training process, and rationale behind the selection of hyperparameters. 
This comprehensive overview enhances the clarity and comprehensibility of our research findings.
\begin{table*}[!t]
\centering
\scriptsize 
\caption{Quantitative comparison in terms of NIQE/BRISQUE (Mean ± Std) on two real-world datasets. Lower scores indicate better image quality. Our method achieves lower NIQE/BRISQUE, resulting in better image quality.}
\renewcommand{\arraystretch}{2}
\setlength{\tabcolsep}{3pt} 
\resizebox{\textwidth}{!}{ 
\begin{tabular}{c|ccccccccc}
\hline
\multirow{2}{*}{Datasets} & PreNet~\cite{12} & MSPFN~\cite{10} & MPRNet~\cite{38.11} & SPANet~\cite{38.05} & Restormer~\cite{36} & MFDNet~\cite{38.14} & MSHFN~\cite{38.13} & Ours \\
\cline{2-9}
 & NIQE/BRISQUE & NIQE/BRISQUE & NIQE/BRISQUE & NIQE/BRISQUE & NIQE/BRISQUE & NIQE/BRISQUE & NIQE/BRISQUE & NIQE/BRISQUE \\
\hline
SPAData~\cite{spadata}   &  4.65±0.32/31.58±0.75  & 4.59±0.34/31.02±0.72 & 4.74±0.31/32.02±0.65 & 4.42±0.28/26.17±0.80 & 5.00±0.30/34.04±0.55 & 4.65±0.33/31.57±0.76 & 4.45±0.29/29.19±0.78 & \textbf{4.35±0.27/26.12±0.82} \\  
MPID~\cite{mpid}   &  3.83±0.35/29.61±0.88  & 3.81±0.33/29.05±0.85 & 3.72±0.36/26.39±0.90 & 3.64±0.30/24.20±0.95 & 3.55±0.28/27.14±0.85 & 3.52±0.32/26.55±0.89 & 3.62±0.33/27.22±0.80 & \textbf{3.31±0.29/23.22±0.92} \\

\hline
\end{tabular}
}
\label{tab:t7}

\end{table*}
\begin{figure*}[!t]
\begin{center}
\begin{tabular}{ccccccc}
\hspace{-1.5mm}\includegraphics[width = 0.14\linewidth]{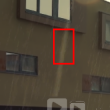} &\hspace{-4.5mm}
\includegraphics[width = 0.14\linewidth]{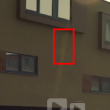}
&\hspace{-4.5mm}
\includegraphics[width = 0.14\linewidth]{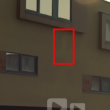} &\hspace{-4.5mm}
\includegraphics[width = 0.14\linewidth]{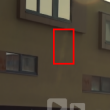} &\hspace{-4.5mm}
\includegraphics[width = 0.14\linewidth]{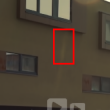} &\hspace{-4.5mm}
\includegraphics[width = 0.14\linewidth]{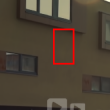} &\hspace{-4.5mm}
\includegraphics[width = 0.14\linewidth]{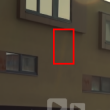} 
\\
\hspace{-1.5mm}\includegraphics[width = 0.14\linewidth]{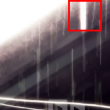} &\hspace{-4.5mm}
\includegraphics[width = 0.14\linewidth]{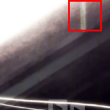} &\hspace{-4.5mm}
\includegraphics[width = 0.14\linewidth]{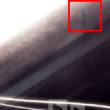} &\hspace{-4.5mm}
\includegraphics[width = 0.14\linewidth]{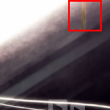} &\hspace{-4.5mm}
\includegraphics[width = 0.14\linewidth]{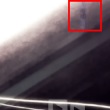} &\hspace{-4.5mm}
\includegraphics[width = 0.14\linewidth]{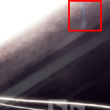} &\hspace{-4.5mm}
\includegraphics[width = 0.14\linewidth]{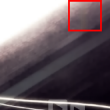}
\\
\hspace{-1.5mm}\includegraphics[width = 0.14\linewidth]{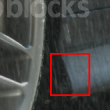} &\hspace{-4.5mm}
\includegraphics[width = 0.14\linewidth]{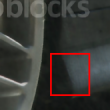} &\hspace{-4.5mm}
\includegraphics[width = 0.14\linewidth]{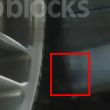} &\hspace{-4.5mm}
\includegraphics[width = 0.14\linewidth]{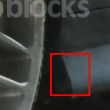} &\hspace{-4.5mm}
\includegraphics[width = 0.14\linewidth]{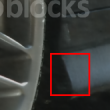} &\hspace{-4.5mm}
\includegraphics[width = 0.14\linewidth]{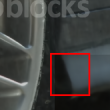} &\hspace{-4.5mm}
\includegraphics[width = 0.14\linewidth]{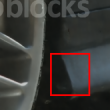}
\\
\hspace{-1.5mm}(a) Rain Image   &\hspace{-4.5mm} (b) SPANet &\hspace{-4.5mm}  (c) Restormer &\hspace{-4.5mm} (d) MFDNet &\hspace{-4.5mm} (e) MSHFN &\hspace{-4.5mm} (f)  NERD&\hspace{-4.5mm} (g) \textbf{Ours}
\end{tabular}
\end{center}

\caption{Visual comparison on the real-world images.
Our EMResformer produces significantly clearer images with sharper details and effectively handles diverse rainy conditions to better eliminate rain streak and restore clarity.}
\label{fig:p12}

\end{figure*}
%
\subsection{Comparison with single image deraining Models on Synthetic Benchmarks}
\subsubsection{Quantitative Results}
We evaluate EMResformer against several advanced single image deraining models.
Tab.~\ref{tab:t1} summarises the results, where our EMResformer achieves superior results, this progress can be much more significant.
Concretely, our method surpasses OHCNet \cite{38.12}, MPRNet \cite{38.11} by $0.319$dB, $0.358$dB on Rain200H, $0.318$dB, $2.911$dB on Rain200L, $0.314$dB, $0.376$dB on Rain1200, $0.319$dB, $0.679$dB on Rain1400 and $0.319$dB, $0.597$dB on Rain12. 
Notably, the improvements on large-scale datasets such as Rain1200 and Rain1400 are particularly significant. These benchmarks contain a wide variety of rain types, including spatially varying rainbands, which pose challenges for most deraining methods. Our superior performance on these datasets indicates that EMResformer is highly effective in handling diverse and complex rain patterns, showcasing its robustness and adaptability to real-world scenarios. This substantial progress underscores the potential of EMResformer in advancing single image deraining research.

\subsubsection{Visual Comparison}
Visual comparisons of EMResformer with other deraining models are shown in Fig.~\ref{fig:p5} and Fig.~\ref{fig:p6}. 
Obviously, our approach excels in restoring finer details and textures, resulting in clearer background images.
%
%
This also highlights the strengths of our EMResformer in preserving image features and fine details. 
%
%
%
The background image restored by other methods exhibits either excessive blurriness or incomplete removal of rain streak.
However, our method accurately restores the background direction, resulting in clearer images.
The achievement of excellent visual effects through more realistic image structures and details is demonstrated by our model.
However, existing single image deraining methods always fail to effectively restore clear structures. In contrast, our approach excels in preserving the texture and edges of the original image, aspects frequently compromised by other methods.
This underscores the robustness and precision of our model in effectively handling complex rain streak and achieving high-quality image restoration, thereby establishing a new benchmark in the field.
%

%
%
\begin{figure*}[!t]\footnotesize
\begin{center}
\begin{tabular}{cccccc}
\hspace{-1.5mm}\includegraphics[width = 0.163\linewidth]{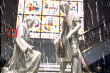} &\hspace{-4.5mm}
\includegraphics[width = 0.163\linewidth]{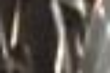}  &\hspace{-4.5mm}
\includegraphics[width = 0.163\linewidth]{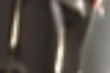} &\hspace{-4.5mm}
\includegraphics[width = 0.163\linewidth]{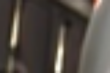} &\hspace{-4.5mm}

\includegraphics[width = 0.163\linewidth]{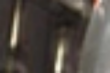} &\hspace{-4.5mm}
\includegraphics[width = 0.163\linewidth]{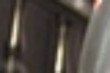} 
\\
\hspace{-1.5mm}(a) Input   &\hspace{-4.5mm} (b) $t=1$ &\hspace{-4.5mm} (c) $t=2$&\hspace{-4.5mm} (d) $t=3$&\hspace{-4.5mm} (e)  $t=4$&\hspace{-4.5mm} (f) GT
\end{tabular}
\end{center}
\vspace{-3mm}
\caption{Visual comparison of different iterations $t$.
The visual effect is best when the number of $t$ is three.
}
\label{fig:p13}
\vspace{-3mm}
\end{figure*}

\subsection{Comparison with Single Image Deraining Models on Real-World Benchmarks}

To evaluate the generalization ability of our model, we perform experiments on the Real-World single image deraining dataset.
As shown in in Tab.~\ref{tab:t7}, our EMResformer get the best results.
Based on average NIQE and BRISQUE scores, our method achieves lower values, indicating superior image quality with greater naturalness and fidelity compared to other models on real-world images.

In Fig.~\ref{fig:p12}, we present the qualitative comparison results, where we can observe that our proposed EMResformer can restore much clearer with sharper details. 

\subsection{Ablation Study}
In this section, we perform ablation studies to examine the impact of various proposed components.
We test the PSNR and SSIM metrics on the Rain200H.
%

%
%
%
\subsubsection{Effect of the EMB}
%
%
We first analyze the effect of EMB within the following experiments.
\begin{itemize}
\item (a) M1: No EMB is used in the model. 
\item (b) M2: The EMB is used in network structure.
\item (c) M3: The IOAB is used in network structure.
\item (d) M4: The IOFN is used in network structure.
\end{itemize}

To evaluate the impact of IOAB and OIFN, we incorporate them into Model M1 to create Model M2.
The method (M1) performs worse when IOAB and IOFN are removed from the encoder-decoder framework.
This proves the effectiveness of the EMB. 
The redundant information in EMB is significantly reduced. We also test the separate IOAB and IOFN modules, M3 and M4, and it turns out that the two modules work best when add to the network structure together. 
Specifically, the redundant information can be reduced better and richer features can be mined. 
From Tab.~\ref{tab:t2}, it can be seen that the performance of the model benefits from IOAB. 
If IOAB and IOFN are added, the PSNR value increase from $29.980$db to $30.221$db.
This improvement is due to IOAB and IOFN’s ability to eliminate redundant information and extract diverse features from the image, which is crucial for effective reconstruction.

To evaluate the impact of IOAB, we visualize the learned features. 
Compared to traditional self-attention, IOAB enhances the reconstruction of detailed features and improves overall restoration quality. 
Due to the optimal iteration attention, we can generate attention maps with less redundant information, thereby allowing the network to focus more on the parts with higher weights, resulting in clearer extraction of rain streak from the rain images.
%
\begin{figure*}[!t]\footnotesize
\begin{center}
\begin{tabular}{cccccc}
\hspace{-1.5mm}\includegraphics[width = 0.163\linewidth]{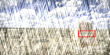} &\hspace{-4.5mm}
\includegraphics[width = 0.163\linewidth]{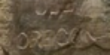}  &\hspace{-4.5mm}
\includegraphics[width = 0.163\linewidth]{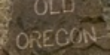} &\hspace{-4.5mm}
\includegraphics[width = 0.163\linewidth]{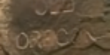} &\hspace{-4.5mm}

\includegraphics[width = 0.163\linewidth]{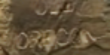} &\hspace{-4.5mm}
\includegraphics[width = 0.163\linewidth]{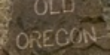} 
\\
\hspace{-1.5mm}(a) Input   &\hspace{-4.5mm} (b) $k=1$ &\hspace{-4.5mm} (c) $k=2$&\hspace{-4.5mm} (d) $k=3$&\hspace{-4.5mm} (e)  $k=4$&\hspace{-4.5mm} (f) GT
\end{tabular}
\end{center}
\vspace{-3mm}
\caption{Visual comparison of different levels of association.
The visual effect is best when the number of cascades is two.
}
\label{fig:p14}
\vspace{-5mm}
\end{figure*}
\begin{table}[t]
\centering
\caption{The comparative results obtained from our method and its variants.
All the results are evaluated on the Y channel.}

\begin{tabular}{c|p{0.5cm}p{0.5cm}p{0.5cm}p{0.5cm}p{0.5cm}p{0.5cm}}
\hline
\multirow{2}{*}{}  & M1         & M2     &M3        & M4       & M5         &M6 
\\
\hline
Base Model & $\checkmark$  &  $\checkmark$  & $\checkmark$ &$\checkmark$ & $\checkmark$ & $\checkmark$ \\ 

IOAB & -  &  $\checkmark$  & $\checkmark$ &- & - & $\checkmark$ \\

IOFN & -  &  $\checkmark$  & - &$\checkmark$ & - & $\checkmark$ \\ 

LMRB & - &  -  & - &- & $\checkmark$ & $\checkmark$ \\ 

PSNR on Rain200H & 29.980 &  30.221  & 29.985 &29.983 & 29.996 & 30.302 \\
\hline
\end{tabular}
\label{tab:t2}
\vspace{-5mm}
\end{table}
\subsubsection{Effect of the Local Model Residual Block}
\begin{itemize}
\item (a) M1: No Local Model Residual Block is used in the model. 
\item (b) M5: The Local Model Residual Block is used in network structure.
\end{itemize}

To evaluate the impact of LMRB, we incorporate it into Model M1, resulting in Model M5.
From Tab.~\ref{tab:t2}, adding LMRB enhances the model's performance, as evidenced by the increase in PSNR from $29.980\text{dB}$ to $29.996\text{dB}$.
This change occurs because the LMRB can help capture short-distance dependencies, thereby enhancing the local modeling ability of the model and helping to reconstruct the details in the image.
\subsubsection{Effectiveness of the iteration t}
We conducted experiments to analyze the influence of the number of iterations $t$, with the results presented in Tab.~\ref{tab:t4}. With the increasing number of $t$ iterations, the performance index of the model is also improving. 
Fig.~\ref{fig:p13} shows the visual effects of different iterations $t$.
More specifically, there is little difference in the performance of the model when $t=3$ and $t=4$ are set. But when $t=2$, compared to the $t=3$ model, the model's performance drops significantly. In real-world scenarios, adjusting $t$ allows us to effectively balance performance and speed.
\begin{table}[t]
\centering
\caption{Performance of models with different $t$ iterations. Test the value of PSNR (Mean ± Std).
All the results are evaluated on the Y channel.}
\begin{tabular}{c|cccc}
\hline
\multirow{2}{*}{}  & Rain200H         & Rain200L     &Rain1200        & Rain1400       
\\
\hline
$t=4$ & 30.299±0.652 & 39.598±0.732 & 34.029±0.850 & 32.930±0.915  \\ 

$t=3$ & \textbf{30.302±0.648} & \textbf{39.600±0.735} & \textbf{34.030±0.849} & \textbf{32.934±0.917}  \\

$t=2$ & 29.985±0.702 & 38.545±0.845 & 33.032±0.911 & 31.845±0.925  \\ 

$t=1$ & 28.548±0.765 & 37.698±0.900 & 32.982±0.960 & 31.585±0.970  \\ 
\hline
\end{tabular}
\label{tab:t4}
\vspace{-4mm}
\end{table}

\subsubsection{Effectiveness of the number of Local Model Residual Block cascades-K}
\begin{figure}[t]
\centering
\includegraphics[width=0.95\linewidth]{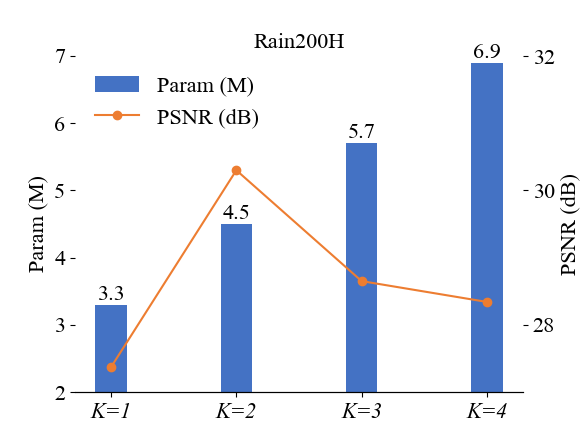}
 \vspace{-3mm}
\caption{Performance of different cascade times $K$. Test the value of PSNR on the Rain200H.
All the results are evaluated on the Y channel.}
\label{fig:p15}
\vspace{-5mm}
\end{figure}

We discuss the influence of different times of parallel connection.
Fig.~\ref{fig:p14} shows the visualized results for the different number of cascades.
The specific experimental results are shown in Fig.~\ref{fig:p15}. 
The model has the best performance when it is cascaded twice. 
A low number of cascades results in insufficient model representation, limiting its ability to fully capture features and model local information effectively. Conversely, an excessive number of iterations reduces the model's generalization ability by increasing overfitting.
\begin{figure}[t]
\includegraphics[width=0.9\linewidth]{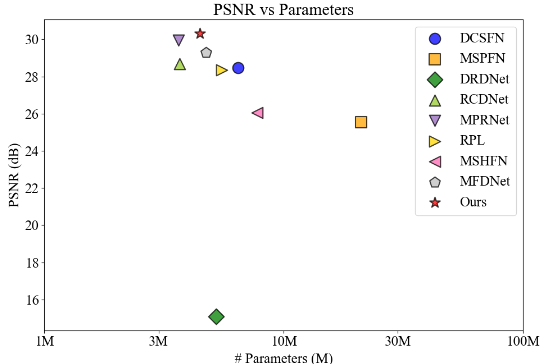}
\vspace{-2mm}
\caption{Comparison between PSNR and Parameters on Rain200H dataset.
All the results are evaluated on the Y channel.}
\label{fig:p15.5}

\end{figure}

\subsubsection{Model Complexity Comparisons}
An analysis of the model’s complexity is essential for a comprehensive evaluation. 
As shown in Fig.~\ref{fig:p15.5}, a comparative study of multi-add operations among various models is presented.
Our EMResformer strikes an optimal balance between model size, performance and computational complexity as measured by multiadds. 
Specifically, our EMResformer ranks first in performance on the Rain200H dataset while boasting fewer model parameters.
When compared with MPRNet \cite{38.11} and MFDNet \cite{38.14}, EMResformer demonstrates superior reconstruction performance, despite having a similar number of parameters and multi-add operations. 
In addition, EMResformer achieves better reconstruction results with fewer model parameters. This makes it highly suitable for implementation in VBMS applications.
%
\begin{figure}[!t]
\centering
\footnotesize 
\setlength{\tabcolsep}{1pt} 
\renewcommand{\arraystretch}{1.0} 
\begin{tabular}{cc}
\includegraphics[width=0.45\linewidth]{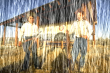} &
\includegraphics[width=0.45\linewidth]{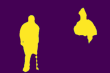} \\
(a)  & (b) MAE = 0.2546  \\
\includegraphics[width=0.45\linewidth]{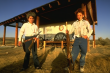} &
\includegraphics[width=0.45\linewidth]{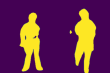} \\
(c)  & (d) MAE = 0.0218  \\
\end{tabular}
\vspace{-1mm}
\caption{Human image segmentation results on the rainy image and the deraining image produced by our method and the Mean Absolute Error (MAE)~\cite{MAE} of saliency map. Our method achieves a smaller MAE value, which implies a better human segmentation effect. (a) Rainy image (b) Rainy image segmentation result (c) Derained image (d) Segmentation result after deraining by our proposed EMResformer.}
\label{fig:p16}
\vspace{-1mm}
\end{figure}

%
\begin{table*}[!t]

\centering
\renewcommand{\arraystretch}{1.5}
\caption{Deraining and Object Detection Results on COCO350/BDD350 Datasets. The \textbf{best} results are marked in bold. Our method achieves the best results.}
\vspace{-1mm}
\begin{tabular}{lccccccc}
\hline
Method & Input & MRPNet~\cite{38.11} & Restormer~\cite{36} & MSHFN~\cite{38.13} & MFDNet~\cite{38.14} & EMResformer (Ours)  \\
\hline
\multicolumn{7}{c}{\textbf{Deraining; Dataset: COCO350/BDD350}} \\
\hline
PSNR  & 14.79/14.14 & 17.99/16.81 & 19.55/18.65 & 18.22/16.91 & 18.38/17.76 & \textbf{20.11/19.33}  \\
SSIM  & 0.649/0.471 & 0.770/0.623 & 0.792/0.685 & 0.775/0.652& 0.783/0.678 & \textbf{0.801/0.694}  \\
\hline
\multicolumn{7}{c}{\textbf{Object Detection; Algorithm: YOLOv5; Dataset: COCO350/BDD350}} \\
\hline
Precision (\%)  & 58.3/47.5 & 65.3/56.4 & 66.2/75.5 & 59.2/72.1& 61.5/74.5  &   \textbf{66.5/76.8}\\
Recall (\%)    & 39.3/44.3 & 56.0/51.0 & 60.1/52.3 & 55.2/33.4 & 59.7/38.7 & \textbf{61.2/55.3}  \\
mAP@0.5 (\%)    & 44.9/32.8 & 61.9/50.6 & 64.5/53.6 & 60.18/50.5 & 63.8/52.2 & \textbf{65.4/54.5}  \\
mAP@.5:.95 (\%) & 26.7/16.3 & 38.6/26.1 & 41.2/30.1 & 35.2/25.8 & 40.0/28.9 & \textbf{42.5/32.6}  \\
\hline
\end{tabular}
\label{tab:t5}
\vspace{-2mm}
\end{table*}
\begin{figure*}[!t]
\centering
\renewcommand{\arraystretch}{1.0} 
\begin{tabular}{ccc}
\includegraphics[width=0.3\linewidth]{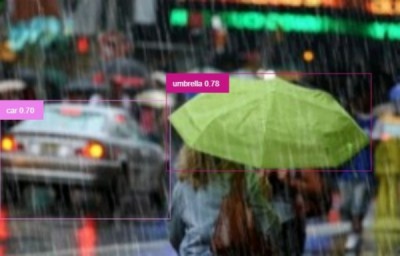} &
\includegraphics[width=0.3\linewidth]{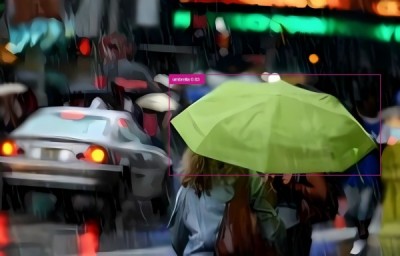} &
\includegraphics[width=0.3\linewidth]{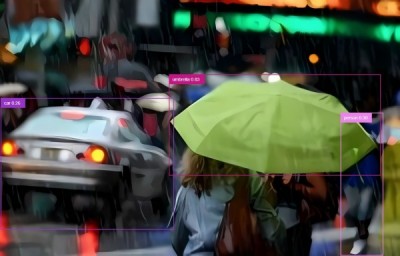} \\
(a) Rain Image  & (b) MPRNet  &  \ (c) Restormer\\
\includegraphics[width=0.3\linewidth]{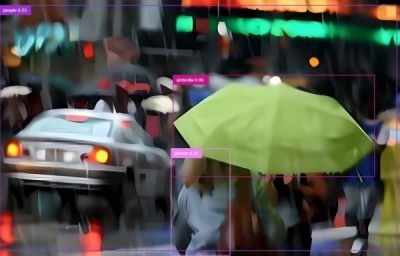} &
\includegraphics[width=0.3\linewidth]{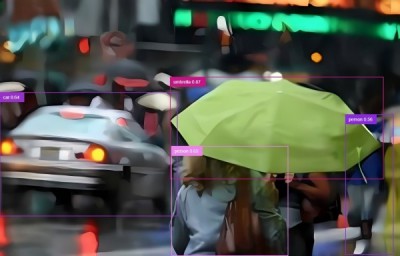} &
\includegraphics[width=0.3\linewidth]{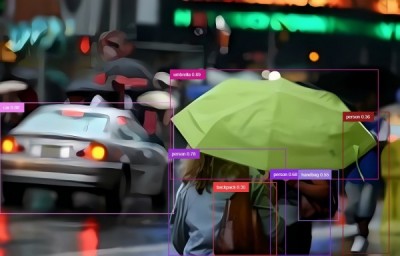} \\
(d) MSHFN  & (e) MFDNet  &  \textbf{(f) Ours}\\
\end{tabular}
\vspace{-1mm}
\caption{Object detection results after deraining using different methods. Our method can better remove rain streak and effectively improve the precision of object detection.}
\label{fig:p555}
\vspace{-5mm}
\end{figure*}
\vspace{-3mm}
\subsection{Limitations}
This paper employs the EM Algorithm to reduce redundant information, thereby enhancing the recovery of fine image details. While our experiments have shown that employing the Expectation Maximization Algorithm improves the recovery of images with sharper details, it also incurs higher computational costs due to the algorithm's requirement for multiple iterations.
\vspace{-3mm}
\subsection{Application in VBMS}
%
%

The advanced vision techniques employed in VBMS are susceptible to imaging degradation, leading to reduced accuracy. To evaluate the effectiveness of our method in enhancing the accuracy of high-level vision tasks in VBMS, we carry out experiments focused on object detection and semantic segmentation.
We use DeepLabV3 to assess the rain streak removal outcomes.
The segmentation outcomes are presented in Fig.~\ref{fig:p16}. The segmented image after the rain has been removed gives the better result. Our EMResformer improves the visual quality of the saliency map and minimizes the Mean Absolute Error (MAE)~\cite{MAE}.
We also use YOLOv5 to assess the impact of rain removal on object detection using the COCO350~\cite{COCO} and BDD350~\cite{BDD}.
Table~\ref{tab:t5} presents the deraining results and the object detection performance after deraining. Our proposed EMResformer achieves the highest PSNR and SSIM values, significantly improves precision and recall, and demonstrates outstanding performance on mAP@0.5 and mAP@.5:.95, highlighting its superiority in image restoration and downstream visual tasks.
Fig.~\ref{fig:p555} presents the qualitative results, clearly showcasing that images processed by our EMResformer outperform others in terms of both rain removal effectiveness and detection accuracy. The enhanced visual clarity and improved object detection capabilities highlight the superior performance of our method compared to existing approaches.

According to the results of our experiment,
preprocessing images to remove rain streak enhances the robustness of segmentation and detection algorithms in advanced vision tasks, leading to more accurate results. 
This enhancement can result in more accurate and reliable measurement outcomes for VBMS.
\section{Conclusion}
This paper introduces the Expectation Maximization Reconstruction Transformer (EMResformer) designed specifically for VBMS through single image rain streak removal, in the context of segmentation and detection.
Noting that self-attention in Transformers can be influenced by irrelevant information, which may lead to redundancy.
We introduce an Iterative Optimal Attention Block to retain the most relevant self-attention values for improved feature aggregation, which is crucial for accurate measurement and analysis in segmentation and detection tasks.
To enhance the extraction of rain features, we design an Iterative Optimal Feedforward Network for more effective multi-scale representation learning.
In addition, a Local Model Residual Block is introduced to provide stronger local model capabilities, thereby maintaining the intricate details in the restored image. This combination effectively addresses the challenges of image quality and detail preservation in measurement contexts.
The experimental results demonstrate that EMResformer outperforms existing advanced methods.
The Expectation Maximization Algorithm can effectively eliminate redundant information, potentially serving as an effective bridge to other measurement and monitoring tasks. 
The results indicate that the proposed EMResformer achieves superior efficiency and stability compared to other state-of-the-art methods in image restoration tasks within VBMS.                                                                   


%

\bibliographystyle{IEEEtran}
\bibliography{sample.bib}
\nocite{40} %
\nocite{41}
\nocite{42}
\nocite{43}
\nocite{44}
\nocite{45}
\nocite{46}
\nocite{47}
\nocite{49}

\vspace{-12mm}
\begin{IEEEbiography}
[{\includegraphics[width=1in,height=1.25in,clip,keepaspectratio]{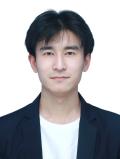}}]{Xiangyu Li} was born in WeiFang, China. He received his B.S. degree in computer science and technology from Dalian University in 2022. Now he is pursuing software engineering in Dalian University and is working hard to pursue a master’s degree. Her research interests include deep learning and computer vision.
\end{IEEEbiography}
\vspace{-12mm}
\begin{IEEEbiography}[{\includegraphics[width=1in,height=1.25in,clip,keepaspectratio]{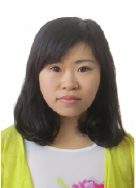}}]{Wanshu Fan} (Member, IEEE) was born in Heilongjiang, China. She received the Ph.D. degree in computational mathematics from the Dalian University of Technology in 2020. Currently, she is working in the School of Software Engineering of Dalian University. Her research interests include computer vision, artificial intelligence, machine learning and big data technology. She is a member of IEEE and CCF.
\end{IEEEbiography}
\vspace{-12mm}
\begin{IEEEbiography}[{\includegraphics[width=1in,height=1.25in,clip,keepaspectratio]{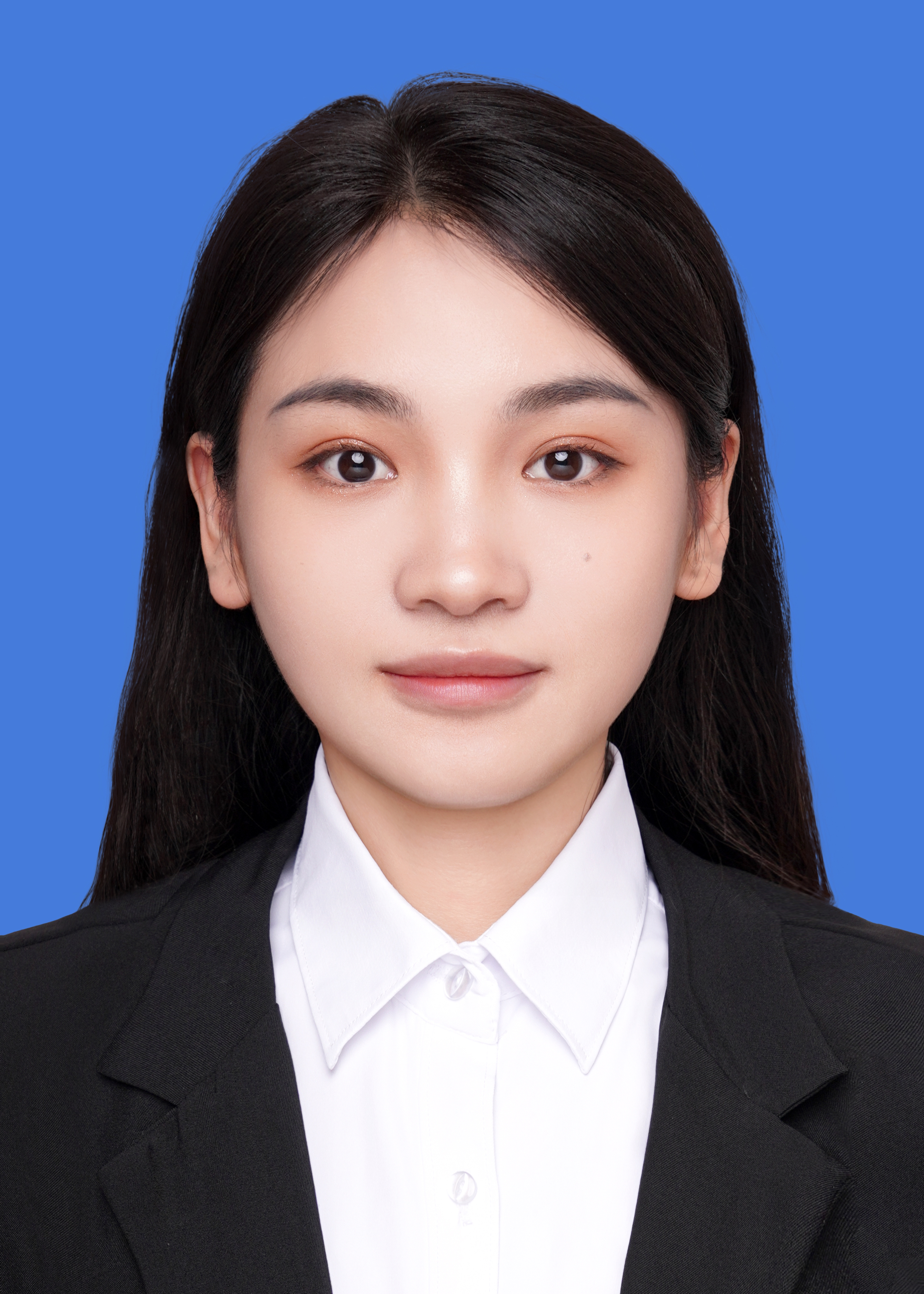}}]{Yue Shen}  was born in Chongqing, China. She received her B.S. degree in software engineering from Chongqing University of Arts and Sciences in 2019. She received the M.S. degree at the school of software engineering from the Dalian University in 2024. Her research interests include deep learning and computer vision.
\end{IEEEbiography}
\vspace{-10mm}
\begin{IEEEbiography}[{\includegraphics[width=1in,height=1.25in,clip,keepaspectratio]{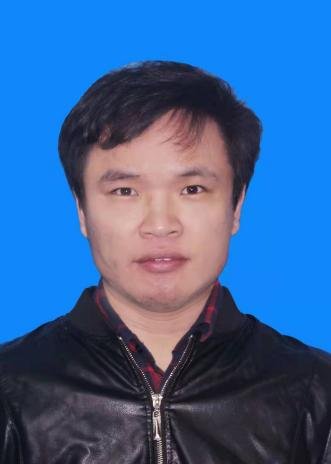}}]{Cong Wang} received the Ph.D. degree at the Department of Computing, The Hong Kong Polytechnic University, in 2024. He received the Master’s Degree in Computational Mathematics from Dalian University of Technology and the Bachelor’s Degree in Mathematics and Applied Mathematics from Inner Mongolia University. His research interests include computer vision and deep learning.
\end{IEEEbiography}
\vspace{-11mm}
\begin{IEEEbiography}[{\includegraphics[width=1in,height=1.25in,clip,keepaspectratio]{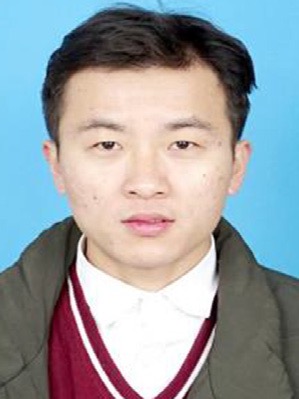}}]{Wei Wang} received the B.S. degree at the school of science from the Anhui Agricultural University, Hefei, China, in 2015. He received the M.S. degree at the School of Computer Science and Technology from the Anhui University, Hefei, China, in 2018. He received the Ph.D. degree with the School of Software Technology, Dalian University of Technology, Dalian, China, in 2022. His major research interests include computer vision, deep learning, etc.
\end{IEEEbiography}
\vspace{-11mm}
\begin{IEEEbiography}
[{\includegraphics[width=1in,height=1.25in,clip,keepaspectratio]{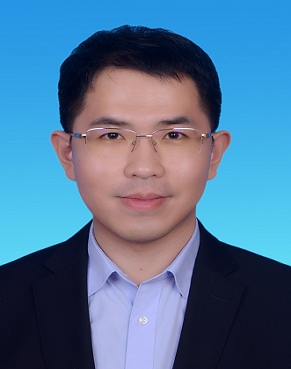}}]{Yang Xin} He is a professor and doctoral advisor at
Dalian University of Technology. He received his Ph.D. degree in computer science from Zhejiang University (2007– 2012) and B.S. degree in com-
puter science from Jilin University (2003–2007). His main research interests include computer graphics and vision, intelligentrobot technology, focusing on
the efficient expression, understanding, perception and interaction of scenes.
\end{IEEEbiography}
\vspace{-11mm}
\begin{IEEEbiography}[{\includegraphics[width=1in,height=1.25in,clip,keepaspectratio]{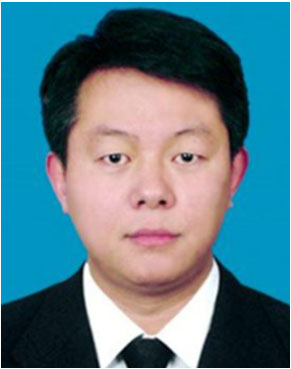}}]{Qiang Zhang} (Senior Member, IEEE) was born in Xian, China, in 1971. He received the M.Eng. degree in economic engineering and the PhD degree in circuits and systems from Xidian University, Xian, China, in 1999 and 2002, respectively. He was a lecturer at the center of advanced design technology, Dalian University, Dalian, China, in 2003 and was a professor in 2005. His research interests are bio-inspired computing and its applications. He has authored more than 200 papers in the above fields. So far he has served in the editorial board of seven international journals and has edited special issues in journals such as Neurocomputing and International Journal of Computer Applications in Technology.
\end{IEEEbiography}
\vspace{-11mm}
\begin{IEEEbiography}[{\includegraphics[width=1in,height=1.25in,clip,keepaspectratio]{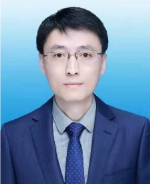}}]{Dongsheng Zhou} (Member, IEEE) received the Ph.D. degree in Dalian University of Technology, in 2010. Currently, he is a Distinguish Professor of Liaoning Province. In 2019, he established the School of Software Engineering of Dalian University and served as the first dean. His research interests include computer graphics, intelligence computing, and human-robot interaction. He was appointed as a member of the Computer Science and Technology Discipline Evaluation Group of the Academic Degrees Committee of the Sixth People’s Government of Liaoning Province. He is a member of IEEE, ACM, CGS and CCF.
\end{IEEEbiography}

\vfill

\end{document}